\documentclass[letterpaper]{article} 
\usepackage{aaai25}  
\usepackage{times}  
\usepackage{helvet}  
\usepackage{courier}  
\usepackage[hyphens]{url}  
\usepackage{graphicx} 
\usepackage{natbib}  
\usepackage{caption} 
\usepackage{algorithm}
\usepackage{algorithmic}
\usepackage{amsmath}

\usepackage{newfloat}
\usepackage{listings}
\DeclareCaptionStyle{ruled}{labelfont=normalfont,labelsep=colon,strut=off} 
\floatstyle{ruled}
\newfloat{listing}{tb}{lst}{}
\floatname{listing}{Listing}
\title{Direct Routing Gradient (DRGrad): A Personalized Information Surgery for Multi-Task Learning (MTL) Recommendations}
\author{
    Yuguang Liu \textsuperscript{\rm 1},
    Yiyun Miao \textsuperscript{\rm 2},
    Luyao Xia \textsuperscript{\rm 3}
}
\affiliations{

    \textsuperscript{\rm 1} Whisper Bond Technologies Inc.\\
    \textsuperscript{\rm 2} Independent Researcher\\
    \textsuperscript{\rm 3} Tongji University\\ 
    log\_whistle@163.com, myothone@gmail.com, luyao.x@tongji.edu.cn
    
}

\begin{document}

\maketitle

\begin{abstract}
Multi-task learning (MTL) has emerged as a successful strategy in industrial-scale recommender systems, offering significant advantages such as capturing diverse users' interests and accurately detecting different behaviors like ``click" or ``dwell time". However, negative transfer and the seesaw phenomenon pose challenges to MTL models due to the complex and often contradictory task correlations in real-world recommendations. To address the problem while making better use of personalized information, we propose a personalized Direct Routing Gradient framework (DRGrad), which consists of three key components: router, updater and personalized gate network. DRGrad judges the stakes between tasks in the training process, which can leverage all valid gradients for the respective task to reduce conflicts. We evaluate the efficiency of DRGrad on complex MTL using a real-world recommendation dataset with 15 billion samples. The results show that DRGrad's superior performance over competing state-of-the-art MTL models, especially in terms of AUC (Area Under the Curve) metrics, indicating that it effectively manages task conflicts in multi-task learning environments without increasing model complexity, while also addressing the deficiencies in noise processing. Moreover, experiments on the public Census-income dataset and Synthetic dataset, have demonstrated the capability of DRGrad in judging and routing the stakes between tasks with varying degrees of correlation and personalization.
\end{abstract}

%

\section{Introduction}

Multi-task learning \cite{Caruana1997}, which leverages information sharing and knowledge transfer between multiple tasks, is widely applied in recommendation systems \cite{c:18}. In real-world recommendation scenarios, different tasks have varying levels of importance. Some tasks, such as "click" or "dwell time," significantly impact online performance and serve as the primary training tasks, despite being challenging to train. The remaining tasks, named “engagement" or “business" heads, can provide finer-grained information and easier to converge, such as " like behavior" reflects the direction and degree of user preference for items. Although these different tasks each have their own emphasis, they are not isolated; instead, they exhibit potential interrelations. Hence, MTL inevitably has the problems of seesaw and negative transfer \cite{c:29}, which means the improvement of a certain task may accompany with others degradation, or some may be affected by the noise of other tasks.

Many existing studies ignore the importance relationship between tasks in business scenarios, such as adaptive weights method \cite{navon2022multi, yang2023adatask} and gradient surgery approaches \cite{c:24, liu2021conflict}. Well hand-crafted MTL Network, like AC-MMOE \cite{Keyao}, solves the seesaw and negative transfer well. Nevertheless, higher model computation will lead to performance issues, especially for the online recommendation \cite{Fabbri2022, c:31, c:32} with stricter response time. IGBv2 \cite{Dai} introduces reinforcement learning to balance the tasks weights dynamically, which is alse computation bound and difficult in convergence. Existing approaches have improved the "seesaw" and "negative transfer" issues, but they also introduce new problems, such as higher model computation, deficiency of noise processing, or ignore the importance among tasks in business.

To address these problems, we utilize the gradient relationship between tasks to concrete stakes between tasks, defined in Fig 1(a), and split the specific task into two parts to reduce its noise impact on the overall loss (shown in Fig 1(b), named Split-MMoE). Motivated by self-supervised router network \cite{Jinyun}, we propose the supervised and end-to-end training router and updater network to strengthen cooperation and reduce conflict. Furthermore, we introduce a personalized gate network, similar to PPNet \cite{hcrt:28}, to mitigate gradient conflicts among users.

\begin{figure}[h]
  \centering
  \includegraphics[width=\linewidth]{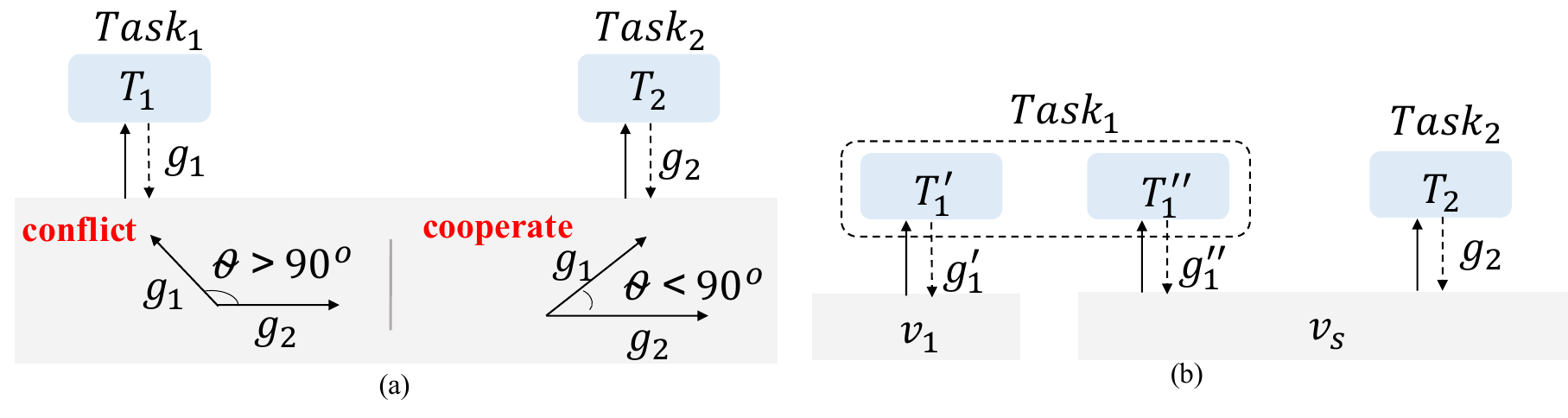}
  \caption{(a) defines $\theta$ as the angle between gradients. When $\theta > 90^\circ$, gradients will update in opposite directions, resulting in conflicts. When  $\theta < 90^\circ$, different gradients will cooperate with each other. (b) seperates $task_1$ into two parts, one uses a dedicated layer and the other shares layer with $task_2$.}
  \vspace{-0.5em}
\end{figure}

In summary, we propose the personalized Direct Routing Gradient (DRGrad) method, which addresses the "seesaw" and "negative transfer" problems without compromising performance or causing information distortion. The main contributions of this work are as follows:
\begin{itemize}
    \item To solve ``seesaw" and ``negative transfer" problems, we propose a router network, which judges the stakes adaptively according to gradient direction between tasks. The router autonomously identifies the optimal gradients from auxiliary tasks and seamlessly integrates them into the current task.
    \item To address the performance and information distortion issues, we adopt a well hand-crafted network structure to divide the task into two parts, assisting the router and updater networks to realize dynamic adjustment. The distinct structures can mitigate the influence of noise on tasks. The updater dynamically aggregates these structures dynamically, guided by the output of the router.
    \item To introduce more personalization information, we propose the personalized gate network. The core method employs a PPNet-like structure \cite{hcrt:28}, using personalized features such as user IDs. Applying this network structure to the underlying share layer of DRGrad can provide personalized gradient information for the router network, which can solve ``seesaw" and ``negative transfer" problems at a finer granularity.
\end{itemize}

\begin{figure*}[h]
  \centering
  \includegraphics[width=0.8\linewidth]{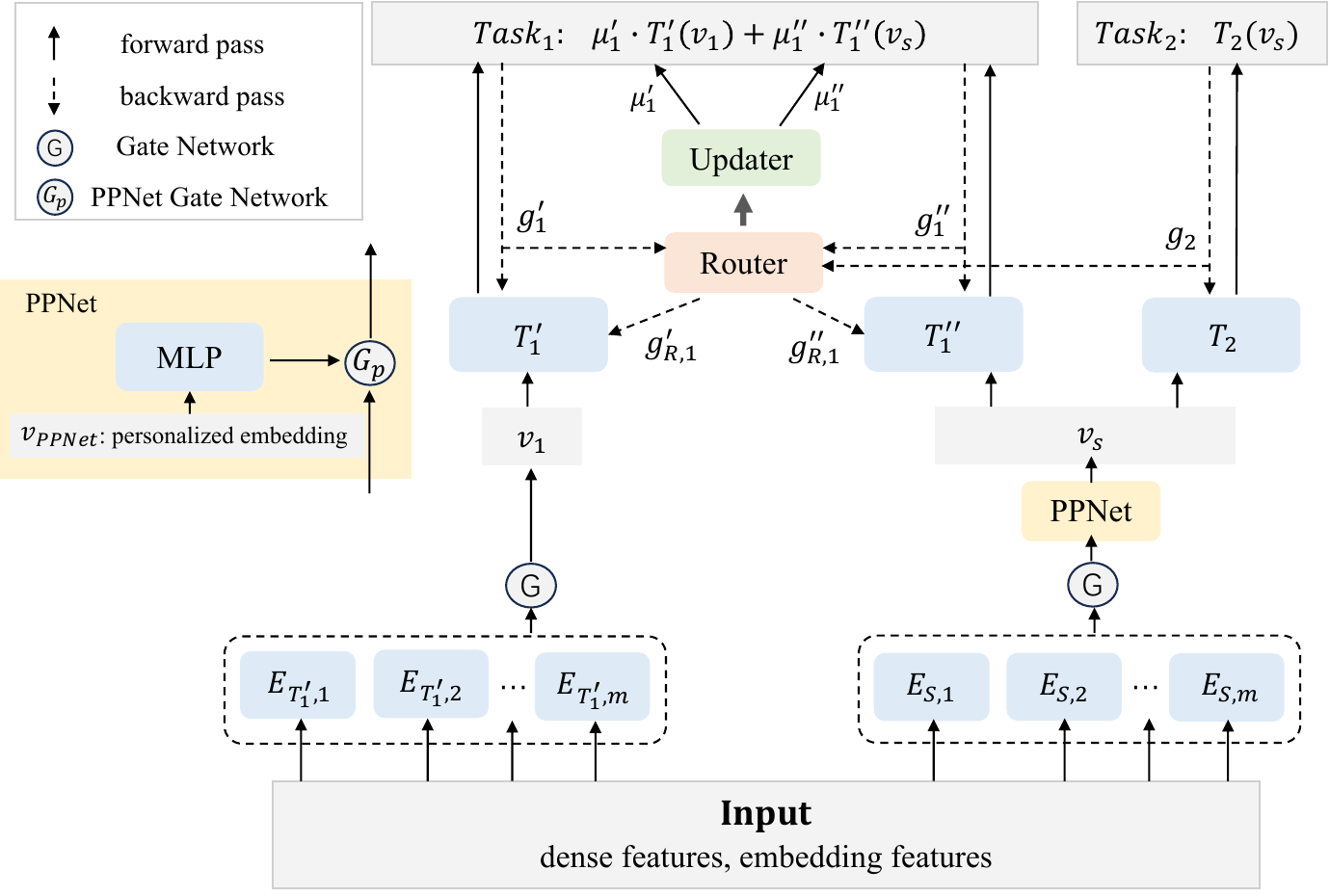}
  \caption{DRGrad model structure. The DNN tower ${T_1}^{'}$ takes the dedicated tensor $v_1$ as its input, and ${T_1}^{''}$ shares the same input tensor, named $v_s$, with ${T_2}$. $Task_1$ is aggregated by the output of ${T_1}^{'}$ and ${T_1}^{''}$, refer as ${T_1}^{'}(v_1)$ and ${T_1}^{''}(v_s)$. The Tensor $v_{PPNet}$ is the input of PPNet, containing the personalized embedding of users. $G$ is the Gate Network, using softmax function and $G_p$ is  Gate Network for PPNet, using sigmoid function.}
\end{figure*}

\section{Related Works}

\subsubsection{Well Hand-Crafted DNN.} MMoE (Multi-gate Mixture-of-Experts) \cite{c:25} implements a gate network for each task to alleviate the conflict. Nevertheless, there is no interaction between experts, which may bring noise and result in the absence of capturing complex information between tasks. Branched MTL \cite{hcrt:22} utilizes employed tasks’ affinities to build branches automatically. SNR \cite{c:26} uses coding variables to control the connection between sub-networks and performs multi-level stacking of the networks, but the dynamic generation is still computation bound. PLE \cite{c:27} improves the efficiency of shared learning and further solves the seesaw from the perspective of joint representation learning, while it's difficult to decouple the complicated relationships. AC-MMoE \cite{Keyao} applies attention and convolution to MMoE to relieve the computation bound. Nevertheless, problem of conflict between tasks still exists in the layers shared by all tasks, and it's difficult for high complexity model to convergent.

\subsubsection{Multi-Task Weight.} The weight or gradient perspective can effectively solve the aforementioned complexity problems. Nash-MTL \cite{navon2022multi} regards gradient combination as a bargaining game, and propose the Nash Bargaining Solution as a principled approach to multi-task learning. IGB \cite{Dai} assigns task weights for improvable gap loss balancing and introduce reinforcement learning to MTL. AdaTask \cite{yang2023adatask} proposes a Task-wise Adaptive learning rate approach and separate the accumulative gradients of each task so that no task would dominate the overall accumulative gradients. It improves the ``seesaw" and ``negative transfer" problems with lower model complexity, but may result in the loss of interactive information. The Pareto optimal solution \cite{c:19} can generate sets of parameters to improve the effectiveness of all indicators, while complex calculations are difficult to implement in industrial scenarios. PCGrad \cite{c:24} defines conflict by the cosine value between gradients direction and rotates the gradients of ``conflict tasks" into the vertical direction. Nevertheless, it may convergent to the Pareto set rather than the optimal point. To solve this, CAGrad \cite{liu2021conflict} seeks the gradient update direction by maximizing the task with the least loss reduction but ignores the overall impact of high-noise tasks. The aforementioned researches tackle the "seesaw" and "negative transfer" problems from the perspective of gradient relationships, providing a less computationally intensive implementation approach, which inspired our work. However, the minimum loss may come from the "engagement" or "business" heads, and continuously optimizing the loss of these tasks may lead to neglecting primary tasks such as "click," disregarding the differences in importance between various tasks in the business, and the impact of the high-noise tasks. Our research work focuses on addressing these issues.

The proposed DRGrad incorporates gradient operations into the model architecture, isolating primary tasks from others, which utilizes router network to maintain the original gradient information. It can route cooperative information to primary tasks without interference from the "engagement" and "business" secondary tasks.

\section{Proposed Method}

The gradient direction of different samples between MTL tasks dynamically changes during training. To leverage cooperative gradients and mitigate conflicts, we propose an end-to-end framework DRGrad, which dynamically judge the stakes between tasks. 

DRGrad comprises of three core components namely Router, Updater and Personalized gate network. The router and updater networks better rectify the gradients to optimize the task's performance, and the personalized gate network is introduced to achieve personalized gradient related to users and fine-grained update of parameters. In training step $t$, we define the relevant quantities:
\begin{itemize}
    \item ${g_1}^{'}(t)$, ${g_1}^{''}(t)$, ${g_2}(t)$: the gradient of ${T_1}^{'}$, ${T_1}^{''}$ and ${T_2}$, where ${T_1}^{'}$, ${T_1}^{''}$ ${T_2}$ are DNN.
     \item $\mu_1^{'}(t)$, $\mu_1^{''}(t)$: the aggregation coefficient from Updater.
\end{itemize}

As illustrated in Fig. 2, $task_1$ is the primary training task, and $task_2$ represent ``engagement" and ``business" heads, which can be expanded to much more tasks. To mitigate conflicts and reduce the impact of noise on the primary task, we partition task1 into two components and introduce the Router network. The Router network routes relevant gradients to the dedicated layer $v_1$ of $task_1$ and differentiates between conflicting and cooperative gradients from the shared layer $v_s$. The updater network collaborates with the router network to dynamically aggregate the two components of $task_1$. The personalized Gate Network employs PPNet to incorporate personalized information into the shared layer, addressing the "seesaw" and "negative transfer" problems at a finer granularity. The code is in appendix A.5

\begin{algorithm}[tb]
\caption{Training Algorithm with DRGrad}
\label{alg:algorithm}
\textbf {Initialize}: $\mu_1^{'}$, $\mu_1^{''}=0.5$, $\gamma > 0$
\begin{algorithmic}[1] 
\FOR{$t=0$ \TO ${max\_train\_step}$}
\STATE Compute $Loss(t)=\Sigma_{i=1}^n \alpha_{task_i}$ (t) * $Loss_{task_i} (t)$
\STATE Compute $g_1^{'}(t)$, $g_1^{''}(t)$,  $g_2(t)$ 
\STATE Compute $g_{R,1}^{'}(t)$, $g_{R,1}^{''}(t)$ by router network (Eq. 1 and Eq. 2)
\STATE Update $\omega_{T_1^{'}}(t)$, $\omega_{T_1^{''}}(t)$ and $\omega_{T_2}(t)$ through Eq. 4 and Eq. 5
\STATE Update all parameters $\omega(t)$ using $\nabla_{\omega_{(t)}}Loss(t)$
\STATE Compute output of updater network $\mu_1^{'}$, $\mu_1^{''}$, through Eq. 6
\ENDFOR
\end{algorithmic}
\end{algorithm}

\subsection{Router Network}
\begin{figure*}[h]
  \centering
  \includegraphics[width=\linewidth]{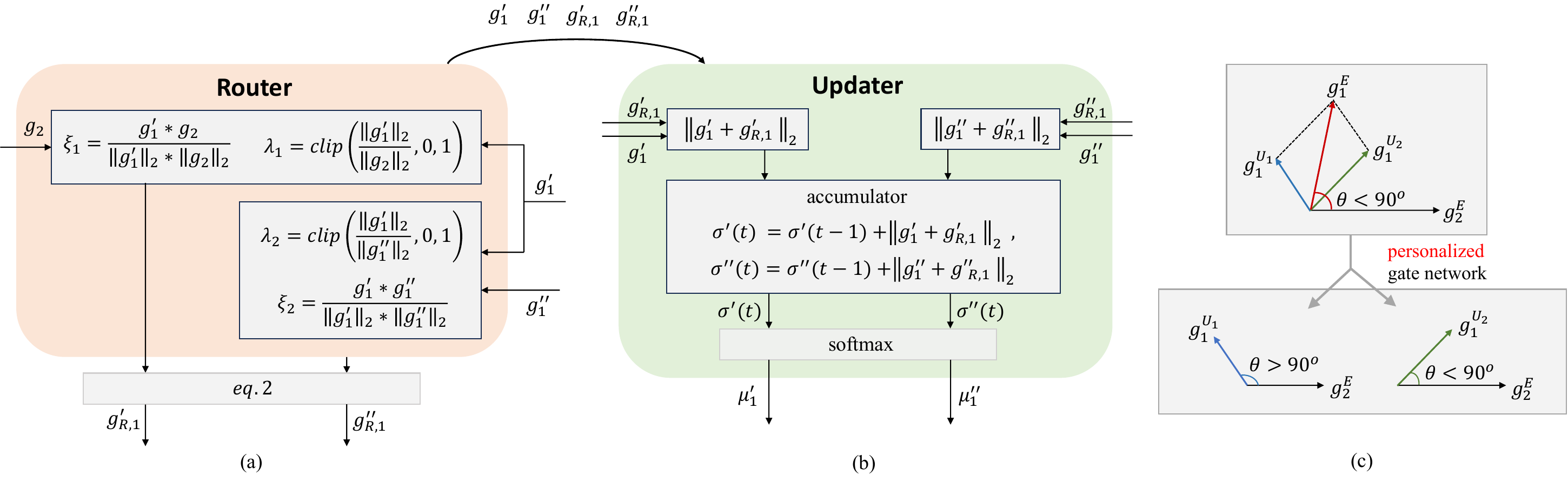}
  \caption{(a) is Router network. The gradients ${g_1}^{'}$, ${g_1}^{''}$ and ${g_2}$ are the inputs of Router Network, which come from $task_1$ and $task_2$. The processed gradients $g_{R,1}^{'}$ and  $g_{R,1}^{''}$ are the outputs, used to update the parameters of ${T_1}^{'}$, ${T_1}^{''}$. (b) is Updater network. Gradient ${g_1}^{'}$, ${g_1}^{''}$, $g_{R,1}^{'}$ and $g_{R,1}^{''}$ are the inputs of Updater Network, and the outputs $\mu_1^{'}$, $\mu_1^{''}$ are used to aggregate $task_1$ dynamically. (c) is Personalized Gradients, $g_1^E$ represents the gradient expectation of all users, $g_1^{U_1}$ represents user $U_1$.}
\end{figure*}

The router network takes effect during back-propagation, which routes the gradient of other tasks to the primary task's DNN network by accessing the relationship among ${g_1}^{'}$, ${g_1}^{''}$ and ${g_2}$. When the $task_1$ is separated by Split-MMoE in Fig. 1(b), the router network will route the coupling information to the corresponding task, while preventing interference between $task_1$ and $task_2$. This approach ultimately improves the accuracy of all tasks.

Fig. 3(a) illustrates the router network, which accesses the influence relationship through the cosine similarity of the gradient. The router network defines similarity $\xi_1$ and $\xi_2$ in Eq. 1, and further calculates the adaptive weights $\lambda_1$ and $\lambda_2$, where ${\Vert \cdot \Vert}_2$ represents the L2 normalization of $x$, and ${\gamma}$ denotes the hyperparameter.
\begin{equation}
\begin{split}
&\xi_1=\frac{g_1^{'} * g_2}{{\Vert g_1^{'}\Vert}_2 * {\Vert g_2\Vert}_2},  \lambda_1\!=\![clip( \frac{\parallel g_1^{'}\parallel_2}{\parallel g_2\parallel_2},0,1)]^{\gamma}\\
&\xi_2=\frac{g_1^{'} * g_1^{''}}{{\Vert g_1^{'}\Vert}_2 * {\Vert g_1^{''}\Vert}_2},  \lambda_2\!=\![clip( \frac{\parallel g_1^{'}\parallel_2}{\parallel g_1^{''}\parallel_2},0,1)]^{\gamma}  \label{1}
\end{split}
\end{equation}

Router's outputs $g_{R,1}^{'}$ and  $g_{R,1}^{''}$, defined in Eq. 2, can provide additional gradient information from $g_1^{''}$ and $g_2$ for ${task_1}$ based on the direction relationship $\xi_1$ and $\xi_2$. $\xi_{i=1,2}$ in the router network determines the value of indicative function $\bf{1}_{\{\textbf{cond}\}}$ in Eq. 3.
\begin{equation}
\begin{split}
& g_{R,1}^{'} \!= \! (1 - {\bf{1}_{\{ \xi_1 < 0 \}}} \! * \! \xi_1) * \lambda_1 \! * \!  g_1^{''} + {\bf{1}_{\{ \xi_2 \geq 0 \}}} \! * \! \lambda_2 \! * \! g_2, \\
& g_{R,1}^{''} \!=\! -{\bf{1}_{ \{ \xi_1 * \xi_2 < 0 \}}} * \xi_1 * \xi_2 * g_1^{''}  
\end{split}
\end{equation}

\begin{equation}
\setlength{\abovedisplayskip}{1pt}
\begin{split}
{\bf{1}_ {\{\textbf{cond}\}}} = 1 \ \ \ \ \ if \ \ \textbf{cond} \ \ else \ \ 0 \label{3}
\end{split}
\setlength{\belowdisplayskip}{1pt}
\end{equation}

As shown in Fig. 2, in training step $t$, the origin gradient $g_1^{'}(t)$ and the output $g_{R,1}^{'}(t)$ of router network are used to update the parameter $\omega_{T_1^{'}}(t)$ of $T_1^{'}$ DNN network in Eq. 4, where $opt$ is the optimizer and $\eta$ denotes learning rate.  
\begin{equation}
\begin{split}
\omega_{T_1^{'}}(t) \leftarrow  \omega_{T_1^{'}}(t-1) - \eta * opt(g_1^{'}(t) + g_{R,1}^{'}(t)) \label{4}
\end{split}
\end{equation} 

For $T_1^{''}$ DNN network, we use its gradient $g_1^{''}(t)$ and router's output $g_{R,1}^{''}(t)$ to update parameter $\omega_{T_1^{''}}(t)$. For $T_2$ DNN network, we use only its gradient $g_2(t)$ to update its parameter.
\begin{equation}
\begin{split}
 & \omega_{T_1^{''}}(t) \leftarrow \omega_{T_1^{''}}(t-1) \! - \! \eta * opt(g_1^{''}(t) \! + \! g_{R,1}^{''}(t)) \label{5}, \\
 & \omega_{T_2}(t) \leftarrow \omega_{T_2}(t-1)  - \eta * opt(g_2(t)) 
\end{split}
\end{equation}

The router network is the superset of PCGrad \cite{c:24}, while its convergence can alse be proven. Router network will not directly rotate the gradient and damage the information, but serve as additional gradient information to promote tasks learning.

\begin{table}[h]
\scalebox{0.92} {
\small
\centering
\begin{tabular}{lc|clc|clc|cl}
\hline
\multicolumn{1}{c}{$\xi_1$} &
\multicolumn{1}{c}{$\xi_2$} &
\multicolumn{1}{c}{Gradient for $v_1$} &
\multicolumn{1}{c}{Gradient for $v_{s}$} \\ 
\hline
\multicolumn{1}{c}{$\geq 0$}  &
\multicolumn{1}{c}{$\geq 0$}  &
\multicolumn{1}{c}{$\! g_1^{'} \! + \! \beta_1 \! * \! g_1^{''}  \! + \!  \beta_2 \! * \! g_2$ \!}  &
\multicolumn{1}{c}{$\! g_2  \! + \!  g_1^{''} \!$}  \\
\multicolumn{1}{c}{$\geq 0$}  &
\multicolumn{1}{c}{$< 0$}  &
\multicolumn{1}{c}{$g_1^{'}  \! + \!  \beta_1 \! * \! g_1^{''}$}  &
\multicolumn{1}{c}{$\! g_2  \! + \!  (1 \! - \!  \xi_1 \! * \!  \xi_2)  \! * \!  g_1^{''} \!$}  \\
\multicolumn{1}{c}{$< 0$}  &
\multicolumn{1}{c}{$\geq 0$}  &
\multicolumn{1}{c}{$g_1^{'}  \! + \!  \beta_1 \! * \! (1 \! - \! \xi_1) \! * \! g_1^{''}  \! + \!  \beta_2 \! * \! g_2$}  &
\multicolumn{1}{c}{$\! g_2  \! + \!  (1 \! - \!  \xi_1 \! * \!  \xi_2)  \! * \!  g_1^{''} \!$}  \\
\multicolumn{1}{c}{$< 0$} &
\multicolumn{1}{c}{$< 0$}  &
\multicolumn{1}{c}{$g_1^{'}  \! + \!  \beta_1 \! * \! (1 \! - \! \xi_1) \! * \! g_1^{''}$}  &
\multicolumn{1}{c}{$\! g_2  \! + \!  g_1^{''} \!$}  \\
\hline
\end{tabular}
}
\caption{Analysis of router network. According to the direction between gradients, the output will be discussed in four cases.}
\label{table1}
\vspace{-1.0em}
\end{table}

Table 1 presents the gradients of the upstream layers operated by the router network, where $\beta_i$ is the coefficient constant. For $v_{s}$, $task_1^{''}$ shares the same vector $v_{s}$ with $task_2$, which facilitates more effective information sharing between the two tasks. The routed gradients $g_{R,1}^{''}$ to $task_1^{''}$ can avoid ``seesaw" between two tasks in this layer. For this layer alone, the router network is equivalent to PCGrad algorithm, however, the key difference is that DRGrad will route the gradient to the downstream DNN parameters of tasks. For $v_1$, the router network can assess information that contributes to $task_1{'}$ from both $task_1{''}$ and $task_2$. In addition, $task_1{'}$ and $task_1{''}$ share the same label, and the rotated $g_1^{''}$ can provide additional task fusion information.

Regarding convergence, the router network performs an incremental operation on the existing gradient. With the clip limitation $0 \leq  \lambda_j  \leq  1$, $0 \leq E[{\bf{1}_{\{X\}}}(\xi_j) * \xi_j] \leq E(\xi_j) \leq 1$, and $0 \leq E[{\bf{1}_{\{X\}}}(\xi_j)] \leq 1$, according to Eq. 8, DRGrad maintains the original gradient direction, and the scale values remain bounded. Consequently, the training process is guaranteed to converge.

\begin{equation}
\setlength{\abovedisplayskip}{1pt}
\begin{split}
g & = g_1^{'} + g_1^{''} + g_2 + g_{R,1}^{'} + g_{R,1}^{''} \\
& = g_1^{'} \! + \!  \! ( \! 1 \! + \! {\bf{1}_{ \! \{ \! X \! \} \! }} \! * \! \lambda_2 \! ) \!  \! * \! g_2 \! + \!  \! ( \! 2 \! - \! {\bf{1}_{ \! \{ \! X \! \} \! }}  \! * \! \lambda_1 \! - \! {\bf{1}_{ \! \{ \! X \! \} \! }} \! * \! \xi \! ) \!   \! * \! g_1^{''} \\
|g| & \leq |g_1^{'}| + |2 * g_2| + |2 * g_1^{''}|
\end{split}
\setlength{\belowdisplayskip}{1pt}
\end{equation}

\subsection{Updater Network}

The updater network is designed to cooperate with the router network to achieve dynamic weight update for task aggregation during each training step $t$. Specifically, to prevent mutual influence between tasks, we divide $task_1$ into two components, which are placed in the dedicated layer and the shared layer respectively. 
As depicted in Fig. 3(b), to dynamically obtain the weights of two components, we employ an updater network, which generates dynamic weights, $\mu_1^{'}$ and $\mu_1^{''}$.  These weights change based on the inputs and outputs of the router network and are used to update the msagnitude between $T_1^{'}(v_1)$ and $T_1^{''}(v_s)$.

The updater network updates itself during back-propagation and takes effect during forward-propagation. In training step $t$, accumulated variables $\sigma^{'}(t)$, $\sigma^{''}(t)$ are updated according to the input and output of the router network, and $\mu_1^{'}(t)$, $\mu_1^{''}(t)$ are obtained by applying the softmax function to $\sigma^{'}(t)$ and $\sigma^{''}(t)$, respectively.
\begin{equation}
\begin{split}
& \sigma^{'} \! (t) \! = \sigma^{'} \! (t \! - \! 1) \! + {\Vert g_1^{'} + g_{R,1}^{'} \Vert}_2 \\
& \sigma^{''} \! (t) \! = \sigma^{''} \! (t \! -1 \!) \! + {\Vert g_1^{''} + g_{R,1}^{''} \Vert}_2 \\
& \mu_1^{'}(t) =  \frac{e^{\sigma^{\!'\!}\!(t)}} {e^{\sigma^{\!'\!}\!(t)} + e^{\sigma^{\!'\!}\!(t)}} \ , \  \ \ \ \mu_1^{\!''\!}(t) =  \frac{e^{\sigma^{\!''\!}\!(t)}} {e^{\sigma^{\!''\!}\!(t)} + e^{\sigma^{\!''\!}\!(t)}} \label{6} 
\end{split}
\end{equation}

The final output of ${task_1}$ is the weighted sum of $T_1^{'}(v_1)$ and $T_1^{''}(v_s)$ in Eq. 7, where $\mu_1^{'}$, $\mu_1^{''}$ are variables updated by the output of the updater network automatically through Eq. 6 above during the training process.
\begin{equation}
\begin{split}
T_1 \! = \! \mu_1^{'} \! * \! T_1^{'}(v_1) \! + \! \mu_1^{''} \! * \! T_1^{''}(v_s) \label{7} 
\end{split}
\end{equation}

In summary,  by utilizing the input and output of the router network as the input of the updater network and accumulating the changes, the weights of each component of $task_1$ can be aggregated dynamically.

\subsection{Personalized Gate Network}

The two tasks share the same vector $v_{s}$, which contains information from all tasks. However, the personalized information in $v_{s}$ is limited. The Personalized Gate Network introduces personalized information to the shared layer, aiming to solve the "seesaw" and "negative transfer" problems at a finer granularity. Gradients mostly represent the expected value of all users, rather than the personalized gradient for a specific user. Therefore, the implement of personalized gate network, a PPNet-like structure, can provide finer-grained personalized information for $v_{s}$. Combined with the router, personalized gate network can achieve personalized gradients. PPNet's input $v_{PPNet}$ is consist of personalized features, such as userId, itemId and authorId. The output of personalized gate network is $v_{s} = 2 * v_{s} \otimes sigmoid(v_{PPNet}*\omega_{PPNet})$.

Multiplying the output of PPNet to the $v_{s}$ can enrich personalized information in network. As shown in Fig. 3(c), $g_1^E$ and $g_2^E$ represent the expected value of gradients. The angle between two gradients is denoted by $\theta$, which represents the relationship between the gradients of all users. But for individual users, the relationship between the gradients of each task may differ from the overall. So PPNet can provide personalized gradients of each user, like $g_1^{U_1}$ and $g_1^{U_2}$, which may have different angles compared to the original gradients. For each user, $g_1^{U_1}$ and $g_1^{U_2}$ are more representative of the relationship between different behaviors and items. When incorporating PPNet to $v_{s}$, it will rotate the gradients $g_2$ and $g_1^{''}$ towards more personalized directions. This enables the router network to obtain finer-grained personalized gradients and provide more accurate routing output.

\section{Experiment}

\subsubsection{Baseline Models.} The backbone is MMoE (Multi-gate Mixture-of-Experts) \cite{c:25} with shared bottom structure, and we choose the following MTL models with different shared network architectures for comparison: SNR \cite{c:26}, PLE \cite{c:27}, AC-MMOE \cite{Keyao}, PCGrad \cite{c:24}, CAGrad \cite{liu2021conflict}, AdaTask \cite{yang2023adatask}, Nash-MTL \cite{navon2022multi}, and IGBv2 \cite{Dai} algorithms, which are the same amount of parameters with DRGrad to verify the effectiveness. Experiment setup is in appendix A.1.

\subsubsection{Evaluation and ablation Studies.} We use the AUC of each task to measure the model's performance and reflect the noise processing capability. In particular, there is a correspondence between the AUC indicator and online effects. For example, the "click" task corresponds to the online CTR effect, in industrial scenarios, even a small improvement in click AUC (e.g. 0.0010) can lead to a significant increase in online CTR (e.g. 0.8\%). Besides, we consider the training time and latency in online serving to reflect the model complexity. To further investigate the effectiveness of key components proposed in the DRGrad model, we design a series of ablation studies. Three variants are considered to simplify DRGrad by: 1) using Split-MMoE network only to validate its effectiveness, 2) using Split-MMoE in collaborate with the router, as shown in Fig. 1(b), to examine the effectiveness of split structure. 3) removing the personalized gate network.

\subsection{Effectiveness Verification}

We verify the effectiveness of the proposed DRGrad using a real-world dataset from a recommender, which consists of 15 billion daily samples collected from a real-world application.

\begin{figure}[h]
      \centering
      \includegraphics[width=\linewidth]{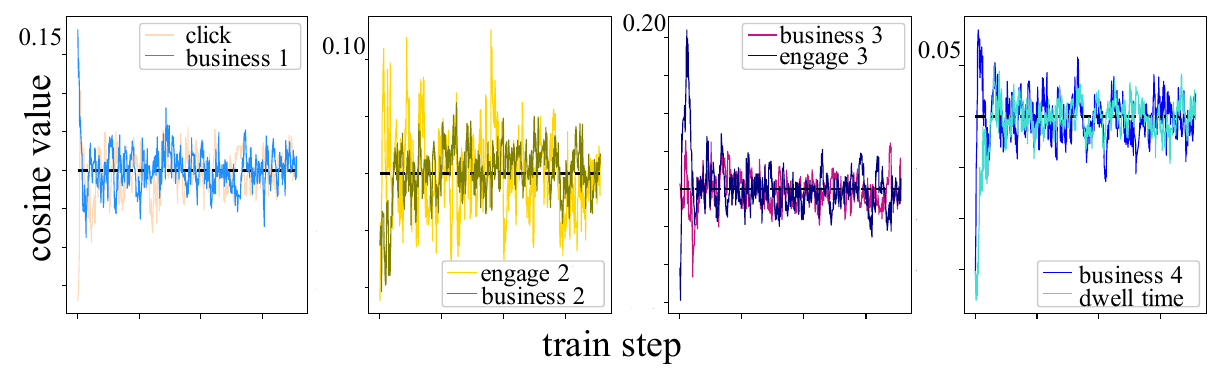}
      \caption{Grad’s direction with respect to click in Fig. 1(b). The gradient direction between tasks fluctuates violently between positive and negative.}
  \hspace{0.1cm}
      \centering
      \includegraphics[width=\linewidth]{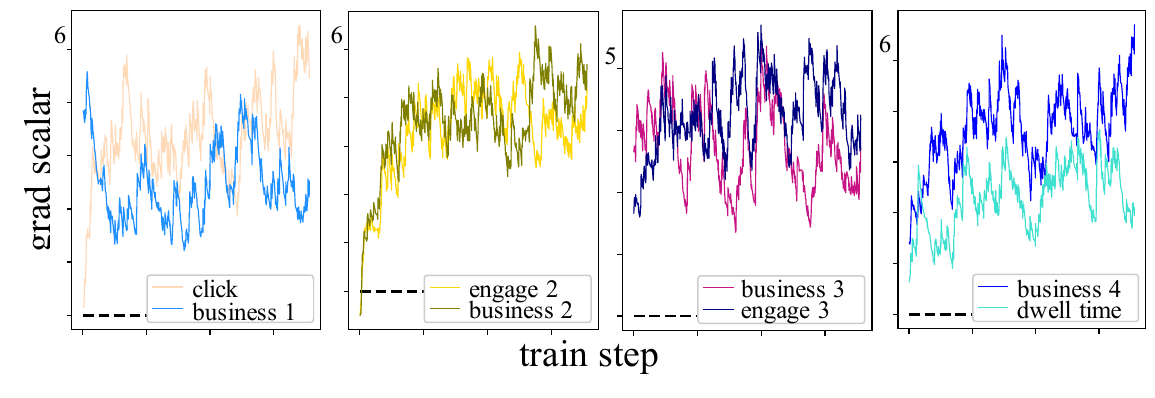}
      \caption{Grad’s scalar with respect to click in Fig. 1(b). The scale of the gradient between tasks is large and the convergence trend is not obvious.}
\end{figure}

\begin{figure}[h]
      \centering
      \includegraphics[width=\linewidth]{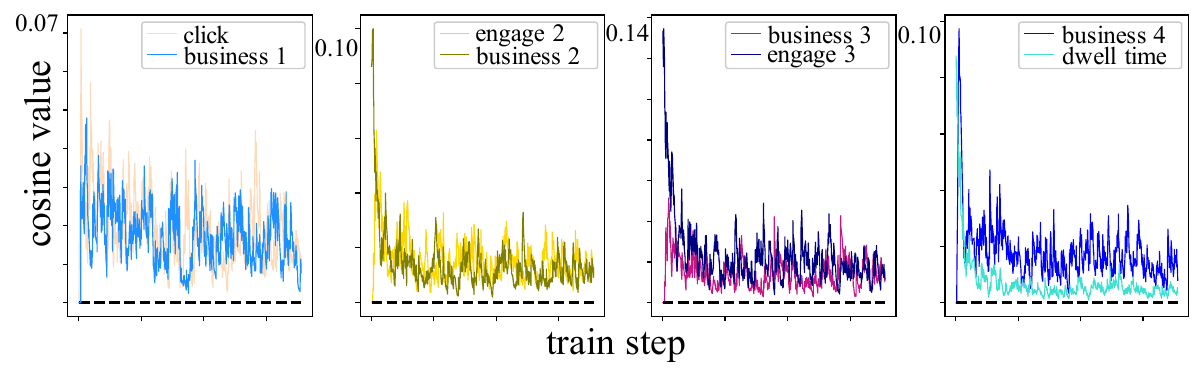}
      \caption{Grad’s direction to click in DRGrad model. The direction between the gradients becomes same direction and is easier to converge.}
  \hspace{0.1cm}
      \centering
      \includegraphics[width=\linewidth]{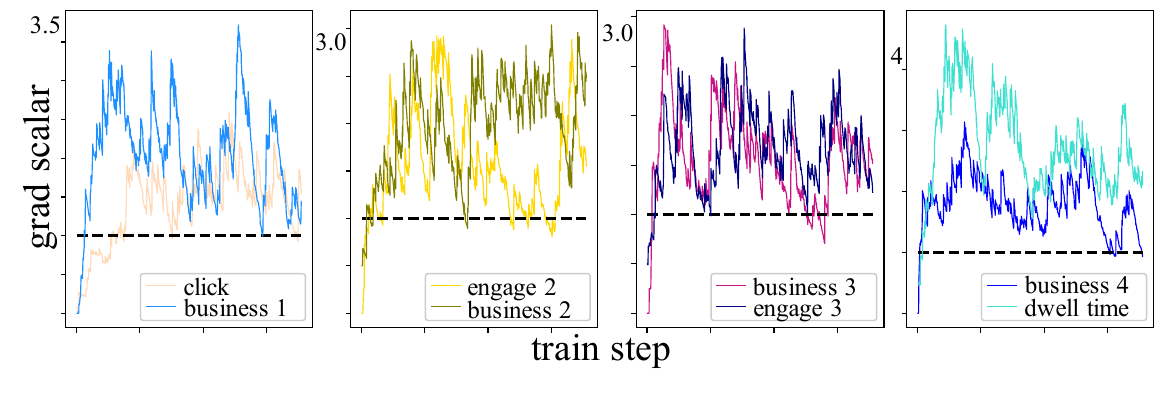}
      \caption{Grad’s scalar to click in DRGrad model. The ratio between gradients becomes smaller and converges faster.}
\end{figure}

\subsubsection{Stakes in Baseline.} The changes of cosine similarity between tasks in baseline model are shown in Fig. 4. During the training process, the cosine values between gradients of ``click" and other tasks fluctuate significantly between positive and negative values. Although the cosine values exhibit a convergence trend, the trend for the "engagement2 task" is not evident. Therefore, each task constantly alternates between conflict and cooperation with the ``click" task, so as to other tasks.  Fig. 5 depicts the variation in gradient scale between tasks in the baseline model. The gradients of each task exhibit large scales and fluctuations, which can affect the gradient updates of the shared layers. In addition, the convergence trend is not apparent during the training process. 

\subsubsection{Effectiveness for DRGrad.} Fig. 6 illustrates the gradient relationships between the primary task and auxiliary tasks in the DRGrad model. With the incorporation of router network, the auxiliary task positively affects the updating of the current primary task. Moreover, compared to the baseline results in Fig. 4, the convergence trend of the cosine values for each task is more pronounced towards zero, which shows that the direction of each gradient relative to the primary task gradually changes to the vertical direction. Thereby, the DRGrad model enhances the cooperation and has certain regularity for the conflict, further simplifying the complex relationships in the shared tower. In DRGrad model, the gradient scales between the primary task and others are shown in Fig. 7. Compared with the baseline results in Fig. 5, DRGrad model has apparent normative effect on the scale of the gradient, which can more effectively prevent task from being affected by other tasks with larger gradient.

\begin{table*}[t]
\small
\centering
\begin{tabular}{lc|clc|clc|clc|}
\hline
\multicolumn{1}{c}{} &
\multicolumn{3}{c}{Cooperate, $E(cos(g_1, g_2))\geq0$} &
\multicolumn{3}{c}{Conflict, $E(cos(g_1, g_2))<0$}   \\ 
\multicolumn{1}{c}{}  &
\multicolumn{1}{c}{MMoE}  &
\multicolumn{1}{c}{Split-MMoE(Fig. 1(b))}  &
\multicolumn{1}{c}{DRGrad} &
\multicolumn{1}{c}{MMoE}  &
\multicolumn{1}{c}{Split-MMoE(Fig. 1(b))}  &
\multicolumn{1}{c}{DRGrad} \\
\hline
\multicolumn{1}{c}{$label_1$ AUC}  &
\multicolumn{1}{c}{0.9521}  &
\multicolumn{1}{c}{0.9568}  &
\multicolumn{1}{c}{\underline{0.9710}} &
\multicolumn{1}{c}{0.8735}  &
\multicolumn{1}{c}{0.8807}  &
\multicolumn{1}{c}{\underline{0.9212}} \\
\multicolumn{1}{c}{$label_2$ AUC}  &
\multicolumn{1}{c}{0.9473}  &
\multicolumn{1}{c}{0.9544}  &
\multicolumn{1}{c}{\underline{0.9596}} &
\multicolumn{1}{c}{0.8712}  &
\multicolumn{1}{c}{0.8828}  &
\multicolumn{1}{c}{\underline{0.9140}} \\
\hline
\end{tabular}
\caption{Comparison of effects on synthesized dataset. Best results are underscored. Regardless the cooperative or conflict relationship between tasks, DRGrad performs better.}
\label{table2}
\vspace{-1.0em}
\end{table*}

\subsection{Artificially Synthesized Dataset Results}

The real-world dataset cannot completely decouple the cooperation and conflict between tasks. To verify the model's effectiveness in reducing conflict and enhancing cooperate between tasks, we designed a synthetic dataset in appendix A.4 with labels indicating absolute conflict or cooperation. The synthesized dataset consists of 110,000 samples, with 100,000 used for training and the remaining 10,000 for testing. The dataset contains 32 features, 6 of which are sparse. $\theta$ is the artificially direction between the task $task_1$ and secondary task $task_2$ while $x$ denotes the input features used to generate labels. The functions $rand(a, b)$, $randint(a, b)$ represent random numbers and random integers between $a$ and $b$, respectively. $N(a, b)$ represents a random value from a normal distribution with mean a and variance b. $label_1$ and $label_2$ are the labels of the primary task $task_1$ and secondary task $task_2$, respectively. To introduce conflict between the two tasks, we set $-1 \! < \! cos(\theta) \! < \! 0$. For cooperation, we set $0 \! < \! cos(\theta) \! < \! 1$.
\begin{itemize}
    \item $i$th sparse feature: $x\!=\!e^{rand(0, 1) * randint(1, i+2)} \!+\! {rand(0, 1) \!*\!randint(1, i\!+\!2)}^{\frac{i}{2}\!+\!1}$
     \item $label_1$: $10 \! * \! (\frac{4 * x^2}{{\Vert x^2 \Vert}_2} \! + \! 5 \! * \! e^{\frac{x}{{\Vert x \Vert}_2}} \! + \! 6*sin(x) \! + \! N(0.01, 0.002)$
     \item $task_2$'s label $label_2$: $cos(\theta) \! * \! label_1 \! + \! N(0.01, 0.002)$
\end{itemize}

Table 2 demonstrates that DRGrad achieves improvements on both tasks, with a more significant improvement on the primary task $label_1$. When the two tasks are in the same direction, DRGrad slightly increases the AUC of $label_2$ and significantly increases the AUC of $label_1$. When the two tasks are in opposite directions, DRGrad yields more substantial improvements in the AUC of both tasks. These results indicate that DRGrad can effectively alleviate conflicts while enhancing cooperation between tasks.

\begin{table*}[t]
\small
\centering
\resizebox{\textwidth}{!}{
\begin{tabular}{lc|clc|clc|cl}
\hline
\multicolumn{1}{c}{} &
\multicolumn{6}{c}{15 Billion Samples Industry Data} &
\multicolumn{2}{c}{UCI Census-Income Data}  \\ 
\multicolumn{1}{c}{Method} &
\multicolumn{1}{c}{Click AUC} &
\multicolumn{1}{c}{Click Gain} &
\multicolumn{1}{c}{Dwell Time AUC}  &
\multicolumn{1}{c}{Dwell Time Gain}  &
\multicolumn{1}{c}{Train Time}  &
\multicolumn{1}{c}{Latency}  &
\multicolumn{1}{c}{Task1 AUC}  &
\multicolumn{1}{c}{Task2 AUC}  & \\
\hline
\multicolumn{1}{c}{MMoE}  &
\multicolumn{1}{c}{0.7624}  &
\multicolumn{1}{c}{-}  &
\multicolumn{1}{c}{0.7477}  &
\multicolumn{1}{c}{-}  &
\multicolumn{1}{c}{\textbf{389min}}  &
\multicolumn{1}{c}{113ms}  &
\multicolumn{1}{c}{0.9387}  &
\multicolumn{1}{c}{0.9927}  & \\
\multicolumn{1}{c}{Split-MMoE(Fig. 1(b))}  &
\multicolumn{1}{c}{0.7626}  &
\multicolumn{1}{c}{0.0002}  &
\multicolumn{1}{c}{0.7481}  &
\multicolumn{1}{c}{0.0004}  &
\multicolumn{1}{c}{\textbf{394min}}  &
\multicolumn{1}{c}{114ms}  &
\multicolumn{1}{c}{0.9393}  &
\multicolumn{1}{c}{0.9928}  & \\
\multicolumn{1}{c}{SNR}  &
\multicolumn{1}{c}{0.7636}  &
\multicolumn{1}{c}{0.0012}  &
\multicolumn{1}{c}{0.7480}  &
\multicolumn{1}{c}{0.0003}  &
\multicolumn{1}{c}{437min}  &
\multicolumn{1}{c}{129ms}  &
\multicolumn{1}{c}{0.9519}  &
\multicolumn{1}{c}{0.9943}  & \\
\multicolumn{1}{c}{PLE}  &
\multicolumn{1}{c}{0.7635}  &
\multicolumn{1}{c}{0.0011}  &
\multicolumn{1}{c}{0.7480}  &
\multicolumn{1}{c}{0.0003}  &
\multicolumn{1}{c}{413min}  &
\multicolumn{1}{c}{117ms}  &
\multicolumn{1}{c}{0.9522}  &
\multicolumn{1}{c}{0.9945}  & \\
\multicolumn{1}{c}{AC-MMoE}  &
\multicolumn{1}{c}{0.7637}  &
\multicolumn{1}{c}{0.0013}  &
\multicolumn{1}{c}{0.7483}  &
\multicolumn{1}{c}{0.0006}  &
\multicolumn{1}{c}{453min}  &
\multicolumn{1}{c}{122ms}  &
\multicolumn{1}{c}{0.9523}  &
\multicolumn{1}{c}{0.9945}  & \\
\hline
\multicolumn{1}{c}{PCGrad}  &
\multicolumn{1}{c}{0.7634}  &
\multicolumn{1}{c}{0.0010}  &
\multicolumn{1}{c}{0.7479}  &
\multicolumn{1}{c}{0.0002}  &
\multicolumn{1}{c}{\textbf{391min}}  &
\multicolumn{1}{c}{113ms}  &
\multicolumn{1}{c}{0.9506}  &
\multicolumn{1}{c}{0.9931}  & \\
\multicolumn{1}{c}{CAGrad}  &
\multicolumn{1}{c}{0.7629}  &
\multicolumn{1}{c}{0.0005}  &
\multicolumn{1}{c}{0.7485}  &
\multicolumn{1}{c}{0.0008}  &
\multicolumn{1}{c}{402min}  &
\multicolumn{1}{c}{113ms}  &
\multicolumn{1}{c}{0.9521}  &
\multicolumn{1}{c}{0.9929}  & \\
\multicolumn{1}{c}{Nash-MTL}  &
\multicolumn{1}{c}{0.7635}  &
\multicolumn{1}{c}{0.0011}  &
\multicolumn{1}{c}{0.7482}  &
\multicolumn{1}{c}{0.0005}  &
\multicolumn{1}{c}{396min}  &
\multicolumn{1}{c}{114ms}  &
\multicolumn{1}{c}{0.9534}  &
\multicolumn{1}{c}{0.9946}  & \\
\multicolumn{1}{c}{Adatask}  &
\multicolumn{1}{c}{0.7640}  &
\multicolumn{1}{c}{0.0016}  &
\multicolumn{1}{c}{0.7483}  &
\multicolumn{1}{c}{0.0006}  &
\multicolumn{1}{c}{390min}  &
\multicolumn{1}{c}{113ms}  &
\multicolumn{1}{c}{0.9532}  &
\multicolumn{1}{c}{0.9947}  & \\
\multicolumn{1}{c}{IGBv2}  &
\multicolumn{1}{c}{0.7643}  &
\multicolumn{1}{c}{0.0019}  &
\multicolumn{1}{c}{0.7482}  &
\multicolumn{1}{c}{0.0005}  &
\multicolumn{1}{c}{426min}  &
\multicolumn{1}{c}{126ms}  &
\multicolumn{1}{c}{0.9529}  &
\multicolumn{1}{c}{0.9948}  & \\
\hline
\multicolumn{1}{c}{DRGrad (ours)}  &
\multicolumn{1}{c}{\underline{0.7651$^*$}}  &
\multicolumn{1}{c}{0.0027}  &
\multicolumn{1}{c}{\underline{0.7493$^*$}}  &
\multicolumn{1}{c}{0.0016}  &
\multicolumn{1}{c}{\textbf{395min}}  &
\multicolumn{1}{c}{113ms}  &
\multicolumn{1}{c}{\underline{0.9550$^*$}}  &
\multicolumn{1}{c}{\underline{0.9949$^*$}}  & \\
\hline
\end{tabular}
}
\caption{Test AUCs on real-world dataset with the best results underscored. A small improvement in click AUC (e.g. 0.0010) can lead to a significant increase in online CTR (e.g. 0.8\%) while DRGrad obtains 0.25\% and 0.12\% absolute AUC gain for click and dwell time. $^*$ indicates the statistical significance for $p\leq0.01$ compared with the best baseline over paired t-test.}
\label{table3}
\vspace{-0.5em}
\end{table*}

\subsection{Real-World Dataset Results}

To evaluate the effectiveness of the proposed method on real-world large-scale datasets, we chose the UCI Census-Income Dataset and a Real-World Recommendation Dataset. This allows for more reliable and easily interpretable results in actual business scenarios.

\subsubsection{UCI Census-Income Dataset.} The UCI census-income dataset is based on 1994 census data and consists of 299,285 demographic records of American adults with 40 fe atures. The tasks aim to predict whether the income exceeds \$50K and whether this person’s marital status is never married. We provide the data processing method in appendix A.3. As shown in Table 3, the split structure has brought improvement in AUC, but DRGrad can improve more significantly. Since there are no personalized features like userid in the Census-income dataset, DRGrad w/o PPNet achieves state-of-the-art AUC on both tasks with absolute improvement gains of 0.0028 and 0.0004, respectively.

\subsubsection{Real-World Recommendation Dataset.} The recommendation dataset consists of 15 billion daily samples from a real-world application. There are two main tasks ``click" and ``dwell time", and several auxiliary tasks like ``business" and ``engagement heads". As shown in Table 3, DRGrad model achieves SOTA offline AUC for two main tasks ``click" and ``dwell time" with improvements of 0.25\% and 0.12\%. These improvements have a significant impact on online dwell time and Click-Through Rate (CTR). It is worth mentioning that in the industry, an offline AUC gain of 0.1\% is considered a substantial improvement and can lead to considerable online gains. Compared with well-handed structure, the perspective of gradient or weight will not increase the complexity of the model itself, resulting in almost no change in online latency. DRGrad also shares this advantage. Since the gradient calculation is introduced in the training process, it will often affect the training time by 6 minutes (vs 389 minutes). Compared with the same effect model, the training time is neutral. We conduct an online experiment which obtains the gain of 0.5712\% for APP online global dwell time and 1.79\% CTR gain for the application's online performance, as shown in Table 4. 

\begin{table}[t]
    \small
    \centering
    \begin{tabular}{lc|clc|}
    \hline
    \multicolumn{1}{c}{} &
    \multicolumn{1}{c}{CTR} &
    \multicolumn{1}{c}{Dwell Time} \\ 
    \hline
    \multicolumn{1}{c}{DRGrad model}  &
    \multicolumn{1}{c}{\underline{1.79\%}$^*$}  &
    \multicolumn{1}{c}{\underline{0.5712\%}$^*$}   \\
    \hline
    \end{tabular}
\caption{Online relative gains compared to MMoE). DRGrad obtains 1.79\% CTR gain and 0.5712\% dwell time gain for the online APP compared with MMoE.}
\label{table4}
\hspace{0.1cm}
    \scalebox{0.9} {
    \small
    \centering
    \begin{tabular}{lc|clc|clc|cl}
    \hline
    \multicolumn{1}{c}{} &
    \multicolumn{1}{c}{Click AUC} &
    \multicolumn{1}{c}{Dwell Time AUC} \\
    \hline
    \multicolumn{1}{c}{MMoE}  &
    \multicolumn{1}{c}{0.7624}  &
    \multicolumn{1}{c}{0.7477}  & \\
    \multicolumn{1}{c}{Split-MMoE}  &
    \multicolumn{1}{c}{0.7626}  &
    \multicolumn{1}{c}{0.7481}  & \\
    \multicolumn{1}{c}{Split-MMoE+router network}  &
    \multicolumn{1}{c}{0.7641}  &
    \multicolumn{1}{c}{0.7492}  \\
    \multicolumn{1}{c}{DRGrad w/o PPNet}  &
    \multicolumn{1}{c}{0.7645}  &
    \multicolumn{1}{c}{0.7491}  \\
    \hline
    \multicolumn{1}{c}{DRGrad}  &
    \multicolumn{1}{c}{\underline{0.7651}}  &
    \multicolumn{1}{c}{\underline{0.7493}}  \\
    \hline
    \end{tabular}
    }
\caption{Results of ablation comparison. Three modules, router, updater, and PPNet structures, are intricately interconnected, resulting in enhanced performance outcomes.}
\label{table5}
\vspace{-1.0em}
\end{table}

Ablation comparison in Table 5 reveals that three key components, router, updater and PPNet network, significantly improve AUC besides split structure. Fig. 8(a) presents the overall loss of the baseline and DRGrad models, demonstrating that the DRGrad model is more conducive to model convergence. The AUCs of each task is shown in  Fig. 8(b). The auxiliary tasks have a positive effect, while the primary task has been dramatically improved. Fig. 8(c) shows that the fine-grained routed information can alleviate the complex convergence problem, leading to lower loss for the primary task.

\begin{figure}[h]
  \centering
  \includegraphics[width=\linewidth]{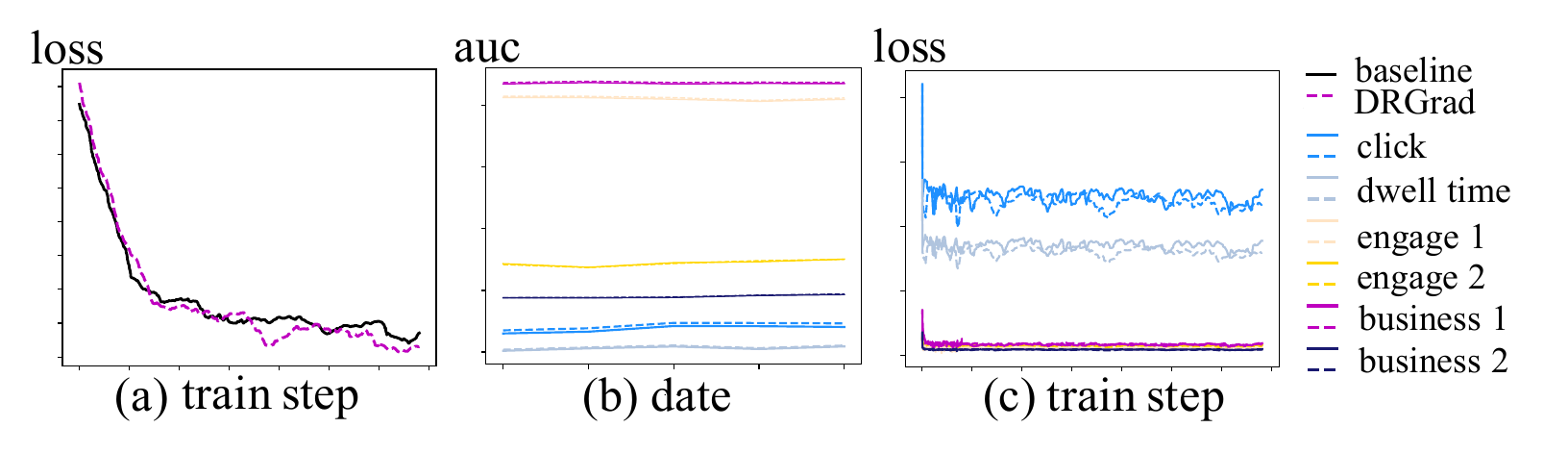}
  \caption{Comparison of loss and AUC (solid line represents baseline, dotted represents DRGrad). DRGrad's loss decreases by an average percent of 3.1 after 300,000 steps.}
\end{figure}

\vspace{-5pt}

\section{Conclusion}

In this paper, we propose Direct Routing Gradient (DRGrad), a novel gradient routing method that effectively mitigates gradient conflicts and enhances the accuracy of Multi-Task Learning (MTL) models. DRGrad incorporates a split model structure and a personalized gate network adapting to the router network, providing regularization and personalization for the intricate information encapsulated within the shared tower. This method leads to better performance on 11 out of 14 tasks in the real-world recommendation system with billions of daily active users and gets better performance on the public Census-income and synthetic dataset compared to MMoE, SNR, PLE, AC-MMOE, PCGrad, CAGrad, AdaTask, Nash-MTL, and IGBv2 algorithms.

\bibliography{aaai25}

\end{document}